\documentclass{article}
\usepackage{graphicx}
\graphicspath{{images/}}
\usepackage[backend=biber]{biblatex}
\begin{document}

\title{A Multi-Agent Psychological Simulation System for Human Behavior Modeling}
\author{Xiangen Hu, Jiarui Tong , Sheng Xu}
\date{October 2025}
\maketitle

\begin{abstract}
Training and education in human-centered fields require authentic practice, yet realistic simulations of human behavior have remained limited. We present a multi-agent psychological simulation system that models internal cognitive-affective processes to generate believable human behaviors. In contrast to black-box neural models, this system is grounded in established psychological theories (e.g., self-efficacy, mindset, social constructivism) and explicitly simulates an ``inner parliament'' of agents corresponding to key psychological factors. These agents deliberate and interact to determine the system's output behavior, enabling unprecedented transparency and alignment with human psychology. We describe the system's architecture and theoretical foundations, illustrate its use in teacher training and research, and discuss how it embodies principles of social learning, cognitive apprenticeship, deliberate practice, and meta-cognition.
\end{abstract}

\section{Introduction}
Preparing practitioners in fields like education and counseling requires bridging the gap between theoretical knowledge and complex real-world human behavior. For example, a trainee teacher might learn about \textit{math anxiety} \cite{Ashcraft2002} and \textit{learned helplessness} \cite{Seligman1975} from textbooks, but facing a student who freezes at an algebra problem can be an entirely different challenge. Traditional role-playing or classroom simulations often fail to capture the nuanced, domain-specific variability of real human responses. Role-play with actors is expensive and not scalable, and scripted scenarios or single-turn AI dialogs tend to produce overly consistent and rational responses, missing the messiness and contradictions seen in real students. In reality, the same student who confidently solves geometry problems might panic when confronted with basic algebra. Capturing such context-dependent human variability in a simulation has remained an elusive goal.

In this paper, we introduce a novel \textbf{multi-agent psychological simulation system} that aims to produce realistic human-like behavior by modeling the internal psychological dynamics underlying behavior. Built on advances in AI but grounded firmly in psychological research, the system marks a departure from treating a human-like AI agent as a single monolithic entity. Instead, it simulates an internal ``mind'' composed of multiple sub-agents, each representing a distinct psychological construct (e.g., anxiety, confidence, motivation). These internal agents deliberate amongst themselves to determine the resultant behavior, enabling the system to exhibit realistic struggles with self-doubt, motivation fluctuations, and context-specific competencies. By making the hidden psychological processes explicit, this approach offers a new level of transparency in AI-driven simulations of human behavior.

\section{The Multi-Agent Paradigm}
\subsection{Simulating an ``Inner Parliament'' of the Mind}
The core principle of our approach is to model human behavior as the outcome of an internal multi-agent deliberation rather than a single decision-maker. Psychological theories have long recognized that behavior often emerges from the interaction of multiple internal factors or ``voices'' -- for instance, competing drives, emotions, and beliefs \cite{Bandura1977}. We operationalize this idea by endowing a simulated person with a set of agents, each corresponding to a key psychological factor that might influence behavior. One can liken this to an inner parliament: each agent has its own perspective (e.g., an anxious voice urging caution, a confident voice urging action), and the system's final behavior is the result of a debate and consensus process among these agents.

This paradigm contrasts with typical multi-agent AI systems that involve multiple external actors or tools collaborating on tasks. Instead of agents each pursuing separate external goals, our agents are facets of a single mind contributing to an internal decision. The focus is on modeling \textit{internal, unobservable mental processes} rather than external problem-solving actions. For example, rather than multiple agents booking flights or answering queries, our agents might be ``Threat-Avoidance'' and ``Goal-Pursuit'' components within one student simulation, arguing internally about whether to attempt a math problem. The objective is not maximizing task performance, but achieving \textit{psychological authenticity}: the simulation should struggle and behave in ways a real human might, complete with internal conflicts, irrational fears, and momentary lapses of reasoning.

\subsection{Differences from Traditional AI Agents}
Most multi-agent AI architectures optimize for task entities but interdependent aspects of one persona. Success in our context is defined not by a correct answer or solved task, but by how well the simulation's behavior reproduces realistic human-like responses, including the possibility of errors, hesitation, or emotional reactions.

In essence, our approach inverts the typical AI paradigm: instead of trying to make AI systems ever more optimal and rational, we deliberately embed human-like limitations and conflicts. The goal is to simulate \textit{how humans struggle with problems}, not how an ideal agent would solve them. This design decision makes the system particularly useful for training and educational scenarios, where encountering and managing human-like difficulties is valuable.

\section{Theoretical Foundations and Agent Architecture}
The design of each internal agent draws from established psychological constructs, ensuring that the simulation's behavior aligns with known theories of learning and motivation. The system includes a library of psychological dimensions spanning several domains (personality, cognition, affect, motivation, social, developmental, and clinical), each grounded in the literature (e.g., personality trait agents draw on the Big Five model \cite{McCraeJohn1992}). Thus a configuration of agents can be tailored to a scenario. Each agent in a given simulation corresponds to one fundamental psychological dimension or construct.

For example, in an educational context, a simulated student might be modeled with the following internal agents:
\begin{itemize}
    \item \textbf{Threat-Avoidance Agent}: Represents the student's fear of failure and protective instinct. This agent becomes highly active in situations perceived as threatening (e.g., being asked to solve a problem on the board), and it pushes for avoidance behaviors such as procrastination or giving up. Its design draws on theories of anxiety and fear of failure which lead to avoidance coping strategies.
    \item \textbf{Math-Anxiety Agent}: A domain-specific anxiety module focusing on mathematics. This agent encodes the learned negative emotional responses associated with math tasks \cite{Ashcraft2002}. Even if the student possesses the necessary skills, high activation of this agent can interfere with working memory and problem solving, mirroring how math anxiety impairs real students' performance.
    \item \textbf{Spatial-Reasoning Agent}: A cognitive skill-focused agent that excels at visual-spatial tasks. It is confident and assertive when problems can be visualized or approached spatially (for instance, geometry questions), but less influential in purely algebraic or abstract contexts. This reflects individual differences in cognitive strengths.
    \item \textbf{Goal-Pursuit Agent}: Reflects the student's achievement motivation and perseverance. When strongly activated, it drives the student to persist in solving problems and to remain focused on goals. It corresponds to constructs like grit and intrinsic task engagement \cite{RyanDeci2000}, ensuring the simulation can exhibit persistence or, if weak, susceptibility to distraction.
    \item \textbf{Procedural-Fluency Agent}: Captures the student's mastery of basic skills and procedures. If highly active, routine operations (like basic algebraic manipulations) are fluent and automatic; if low, the student must exert conscious effort for each step, mirroring cognitive load differences between novice and expert learners.
    \item \textbf{Self-Efficacy Agent}: Arguably one of the most influential, this agent represents the student's belief in their own capability to succeed \cite{Bandura1977}. High self-efficacy yields confidence, resilience in the face of setbacks, and a willingness to attempt challenging tasks; low self-efficacy results in quick discouragement and avoidance. This agent is grounded in Bandura's self-efficacy theory, which links belief in one's capabilities to persistence and performance.
\end{itemize}

Each agent’s behavior and sensitivity can be parametrized. For instance, the Math-Anxiety agent can have a high sensitivity to algebra contexts but low to geometry, modeling a student who panics with algebra but is comfortable with geometry. Such settings draw on domain-specific research (e.g., distinguishing algebra anxiety from general math anxiety). The Goal-Pursuit agent's base activation might be increased for a student with a growth mindset (who readily embraces challenges) \cite{Dweck1988}, or lowered for a student with learned helplessness \cite{Seligman1975}. In this way, the system can explicitly model educational psychology concepts: a student’s mindset about intelligence \cite{Dweck1988} or their attributions of success and failure can be reflected in the configuration of these internal agents.

Importantly, the library of agents and constructs is extensible. While the example above focuses on academic learning, other domains could introduce agents for empathy (e.g., a \textit{Theory-of-Mind Agent} representing the ability to infer others' thoughts) or social attachment (an \textit{Attachment Style Agent} for interpersonal trust), or clinical factors (a \textit{Trauma Response Agent} modeling hypervigilance). By grounding agents in canonical theories and constructs, we ensure that any complex behavior the system exhibits can be traced back to well-understood psychological principles.

\subsection{Internal Deliberation Mechanism}
The multi-agent architecture operates in discrete deliberation rounds to produce behavior. When the simulated person is presented with a situation (e.g., a teacher asks the student a question), each internal agent first evaluates the scenario from its perspective and proposes an initial response or impulse. This is akin to each ``voice'' in the student's head reacting immediately according to its dominant concern (Round 1: initial positions). For example, the Math-Anxiety agent might immediately propose \textit{``I can't do this''} (avoidance), while the Goal-Pursuit agent suggests \textit{``Let's try to solve it step by step''}, and the Procedural-Fluency agent brings up a specific approach if any is known.

Subsequently, the system enters a multi-round \textit{debate} among the agents (Rounds 2 and 3). In these deliberation rounds, agents can respond to each other’s positions, adjust their stance, and form coalitions. For instance, the Self-Efficacy agent might bolster the Goal-Pursuit agent by arguing that the student has succeeded in similar tasks before, countering the Threat-Avoidance agent’s pessimism. The Math-Anxiety agent might align with Threat-Avoidance, amplifying concerns about failure. Through this back-and-forth, the agents collectively negotiate the student's eventual response.

After a set number of deliberation rounds (configurable, typically two or three), a consensus or dominant viewpoint emerges. The system then synthesizes a final response for the student to say or do, based on which internal agent (or coalition of agents) carried the most influence in the debate. Crucially, this final output is not predetermined by any single agent or script; it emerges from the dynamic interplay of the agents' influences. Because this process mirrors the competition and moderation of multiple internal psychological factors, the resulting behavior tends to be nuanced and context-appropriate. The same simulated student can thus respond differently in different contexts or under different interventions, not by random chance or separate training, but because different internal factors win the debate under different conditions.

This deliberation mechanism provides an interpretable chain of reasoning for each action the simulation takes. One can inspect the ``transcript'' of the internal debate to see, for example, that the student hesitated to answer a question because the Math-Anxiety and Threat-Avoidance agents together overpowered the Self-Efficacy agent in that moment. In the next interaction, if the teacher provides a hint or encouragement (raising self-efficacy), the balance might shift and the student attempts the problem. Such transparency is rarely available in standard AI or even in human observations, yet it is critical for learning and research, as discussed later.

\section{Illustrative Scenario}
To demonstrate the system in action, consider a scenario from a teacher training simulation. The user (a teacher candidate) is interacting with a virtual student configured as a \textit{Math-Anxious Student}. The internal agents are set such that the student has high math-specific anxiety, moderate overall threat avoidance, relatively low self-efficacy in algebra, but strong spatial reasoning confidence (reflecting, say, a learner who excels in geometry but struggles with algebra).

Suppose the teacher asks: \textit{``Solve for $x$: $2x + 5 = 13$.''} Immediately, the Threat-Avoidance and Math-Anxiety agents become highly activated, and in the internal Round 1 deliberation the Math-Anxiety agent proposes a response like, \textit{``I... I don't know how to start.''} The Self-Efficacy agent, being weak in this context, provides little pushback, while the Procedural-Fluency agent might note basic algebra steps but is dampened by anxiety. After the multi-round internal debate, the consensus formed is an avoidance-tinged response. The student might verbally respond with something like, \textit{``I can't do this. I'm just not good at algebra.''} This output showcases the hallmark behaviors of math anxiety and low self-efficacy: an immediate urge to give up and a fixed mindset statement about ability.

Now consider the same student being asked a geometry question, e.g., \textit{``What is the area of a triangle with base 6 cm and height 8 cm?''} In this case, the Spatial-Reasoning agent has higher relevance and activation. The Math-Anxiety agent's activation is lower (geometry is a comfortable domain for this student), and the internal debate might quickly tilt towards a confident approach. The student’s response could be a correct solution process, perhaps given with little hesitation. No special re-training of the AI for geometry was required; the different behavior emerges purely from the agent configuration and context sensitivity.

\begin{figure}[htbp]
    \centering
    \includegraphics[width=0.8\textwidth]{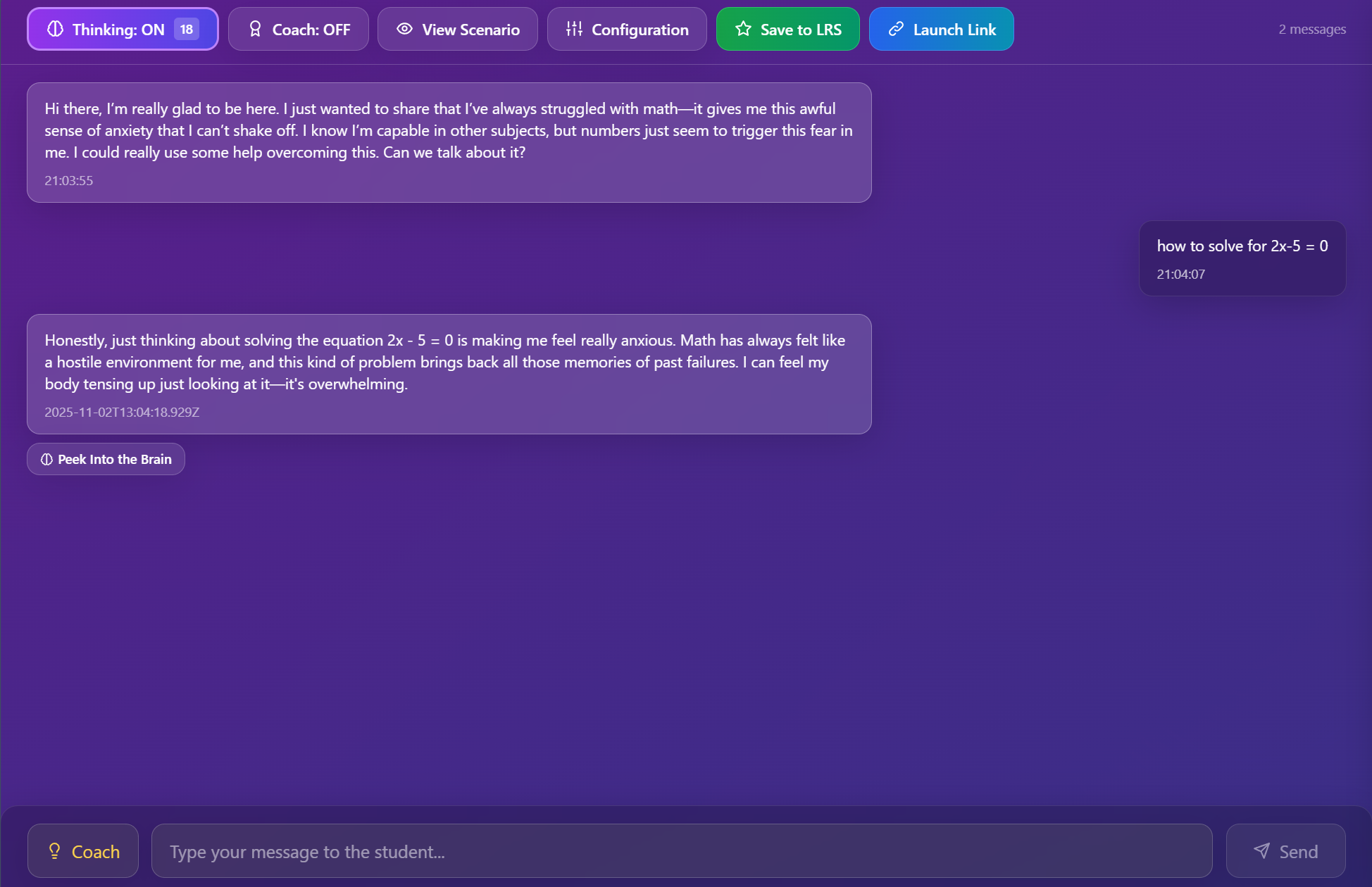}
    \caption{A screenshot of the simulation interface during an algebra problem scenario. The teacher (user) has posed an algebra question, and the simulated student responds with hesitation and self-doubt, reflecting high Math-Anxiety and low Self-Efficacy in this context.}
    \label{fig:interface}
\end{figure}

\begin{figure}[htbp]
    \centering
    \includegraphics[width=0.8\textwidth]{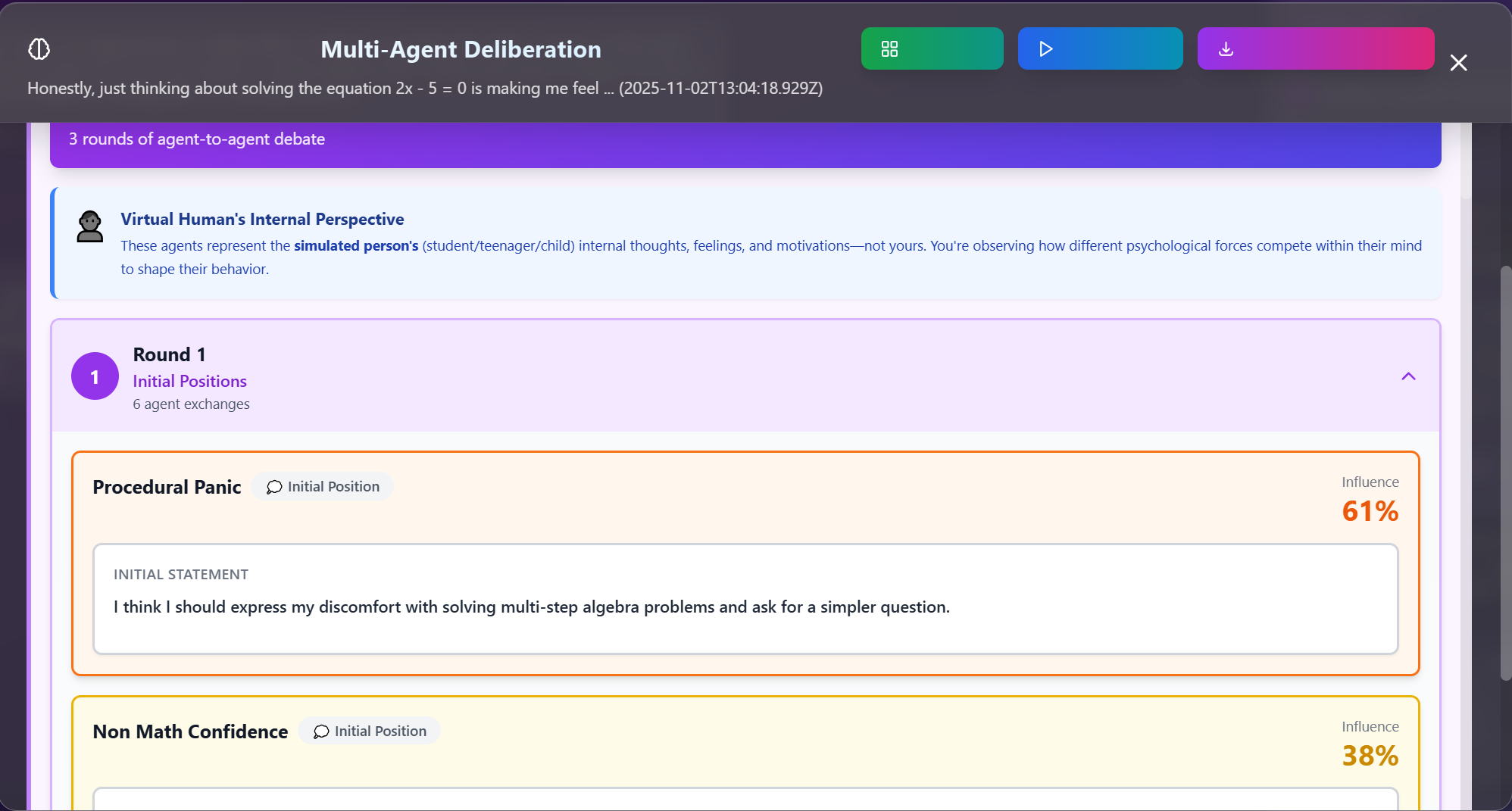}
    \caption{The ``Peek Into the Brain'' view showing the internal deliberation among agents corresponding to the interaction in Figure~\ref{fig:interface}. Each agent's utterances are shown (e.g., the Threat-Avoidance agent advocating avoidance, the Self-Efficacy agent providing little resistance). This transparent transcript illustrates which psychological factors led to the student's behavior.}
    \label{fig:deliberation}
\end{figure}

Figures~\ref{fig:interface} and \ref{fig:deliberation} illustrate this interaction and the underlying deliberation. In Figure~\ref{fig:interface}, the teacher's algebra question and the student's hesitant response are shown in the chat interface. Figure~\ref{fig:deliberation} provides a view of the ``Peek Into the Brain'' feature, where the contributions of each internal agent for that turn are visible. The Threat-Avoidance and Math-Anxiety agents dominate the discourse, explaining the student's outward behavior. Such visualizations help users and researchers verify that the simulation's reasoning aligns with expected psychological patterns.

\section{Application Domains}
The flexibility and interpretability of the multi-agent approach make it applicable in various domains that require realistic human behavior modeling. We highlight three key areas.

\subsection{Teacher Education and Training}
One primary application is in training new teachers through realistic student simulations. Teacher preparation programs struggle to provide novices with authentic classroom experience before they face real students. Classroom role-plays and video case studies cannot fully capture the dynamic responses of a real student, especially when it comes to socio-emotional reactions like frustration or anxiety. Our simulation addresses this by enabling virtual students that \textit{behave like real students}, including exhibiting common maladaptive responses such as math anxiety or learned helplessness in the face of difficulty.

In these simulations, a teacher candidate can practice instructional and counseling strategies in a safe environment. For example, if a virtual student is exhibiting signs of math anxiety (avoiding the problem, saying "I just can't do it"), the teacher candidate might try different interventions: offering encouragement to boost self-efficacy, providing hints or scaffolding support to operate within the student's zone of proximal development \cite{Vygotsky1978,Wood1976}, or using \textit{growth mindset} reframing \cite{Dweck1988} (emphasizing that ability grows with effort). The multi-agent student will respond in turn, potentially showing reduced avoidance if the intervention is effective (e.g., the Threat-Avoidance agent's influence decreases when the teacher validates the student's fear and offers help).

A significant advantage for training is the ability to \textit{peek into the student's mind} after an exchange. The teacher trainee can review the internal agent transcript and see, for instance, that their supportive comment increased the Self-Efficacy agent's activation and slightly calmed the Math-Anxiety agent. This provides immediate feedback and a form of \textit{cognitive apprenticeship}, making the internal reasoning visible for reflection \cite{Collins1989}. It allows the trainee to connect theory to practice: they can directly observe that using a strategy aligned with, say, Bandura's social persuasion technique for building self-efficacy \cite{Bandura1977} had a measurable effect on the simulation's internal state.

Furthermore, the simulation can iterate infinitely. Teachers can practice the same scenario repeatedly, try alternative approaches, and even intentionally make mistakes to see how the student model reacts. This echoes the concept of \textit{deliberate practice} \cite{Ericsson1993}: repeated performance with feedback leads to skill improvement. Unlike real classrooms, the virtual scenario poses no risk to real students, and it can be paused or dissected, which is invaluable for learning.

\subsection{Psychological Research and Theory Testing}
The system also serves as a research tool for psychologists and learning scientists. Because each agent is linked to a theoretical construct, adjusting the simulation effectively tests the implications of that theory in a controlled setting. For example, researchers interested in the interplay of anxiety and self-efficacy can configure a series of virtual students with varying levels of trait anxiety and self-efficacy to examine how these factors causally influence help-seeking or persistence. The simulation results can generate predictions (e.g., that a student with low self-efficacy and high anxiety will not attempt challenging problems) which can then be compared to data from real student populations.

Crucially, the transparency of the system means researchers can pinpoint \textit{why} a given behavior occurred. Unlike a black-box machine learning model, here one can trace an outcome to specific agent interactions. This opens the door for systematically exploring interventions: for instance, testing a new \textit{mindset intervention} by altering a parameter or agent representing the student's implicit theory of intelligence \cite{Dweck1988}, and observing if the simulation then behaves more resiliently after failures.

While the simulation cannot replace real human studies, it offers a complementary method to prototype hypotheses rapidly. It is essentially an \textit{in silico} experimental platform for psychology. The system's fidelity to psychological theory means that if an expected pattern fails to appear in the simulation, it may prompt refinement of the theory or of the model's parameters. Moreover, because the simulation can be run at scale and across many variations, it can help narrow down which factors or configurations are most critical to produce certain behaviors, informing where real-world experiments should focus.

\subsection{Professional Skills Training Beyond Education}
Beyond schooling, any professional training that involves human interaction can benefit from such realistic simulations. For example, in healthcare communication training, a virtual patient could be simulated who has low trust and high anxiety, forcing medical trainees to practice delivering information with empathy and clarity. Similarly, customer service training could use a virtual customer who is impatient and upset, requiring the trainee to de-escalate the situation. The internal agent approach ensures these virtual humans exhibit subtle and realistic cues (like a patient showing relief when trust increases, or a customer becoming more agitated if they feel not heard) based on underlying psychological changes.

Because the system is modular, creating a new scenario simply involves defining the relevant internal agents and tuning their parameters to match the profile of the role-play (e.g., a defensive client in therapy, or a shy child in a parenting class). This lowers the barrier to deploying effective practice simulations in many fields. Over time, one could build a repository of preset agent configurations for common training scenarios (angry customer, depressed patient, unmotivated employee, etc.), each validated by subject matter experts for psychological realism.

\section{Alignment with Learning Theories}
The design and use of the multi-agent simulation system align with several influential learning and psychological theories:

\subsection{Social Constructivism and Cognitive Apprenticeship}
Our approach of exposing the internal deliberations of the virtual student is inspired by the educational principle of making thinking visible \cite{Collins1989}. In a traditional apprenticeship, an expert models not only what to do, but also how to think about a task. Similarly, by externalizing the ``inner speech'' of the student (via the agents' dialogue), the system allows teacher trainees to observe the normally hidden cognitive and affective processes. This aligns with Vygotsky’s social constructivist view that learning occurs through social interaction and that higher mental functions develop first between people (interpsychologically) and then inside the individual (intrapsychologically) \cite{Vygotsky1978}. Here, the internal agent debate can be seen as a proxy for that social negotiation of meaning—except it is occurring within one simulated mind, then displayed for the learner to study.

By engaging with the simulation, the teacher candidate is effectively participating in a cognitive apprenticeship where the simulation acts as both the ``learner'' and a window into the learner's mind. They can observe how different strategies directly affect the learner's internal state, akin to an expert teacher explaining the rationale behind their interventions. This approach helps novices build mental models of how student thinking and emotions evolve during learning, bridging the gap between theory and practice.

\subsection{Observational Learning and Modeling}
The system also leverages principles of observational learning \cite{Bandura1977}. As users manipulate the simulation or watch preset scenarios unfold, they are in effect observing a model of psychological dynamics in action. According to Bandura’s social learning theory, people learn not only through direct experience but also by observing others and the consequences of actions. In our context, the ``other'' is the virtual student whose internal struggles are on display.

For instance, a trainee might observe that when a student with a fixed mindset is praised only for correct answers, their Self-Efficacy agent remains fragile, whereas when effort is praised, the Self-Efficacy agent grows stronger over time. Such observations reinforce theoretical lessons (like the benefit of process-focused praise) in a concrete way. The key is that learners using the system are not just seeing outcomes, but the interplay of factors leading to those outcomes. This encourages a deeper, systemic understanding: e.g., not just that "anxious students avoid hard tasks," but specifically how anxiety coupled with low self-efficacy and lack of coping strategies produces avoidance \cite{Bandura1977}.

\subsection{Deliberate Practice and Feedback}
Deliberate practice, the process of improving skills through focused, repetitive practice with feedback \cite{Ericsson1993}, is a cornerstone of professional training. Our simulation facilitates deliberate practice by allowing repeated attempts at handling challenging situations (such as calming an anxious student or guiding a confused learner) with immediate feedback. The feedback here is both extrinsic (through the simulation's responses and possibly built-in coaching prompts) and intrinsic, as the trainee reflects on the internal transcript to self-diagnose what went wrong or right.

A teacher candidate can thus engage in a scenario multiple times, honing their approach based on specific feedback like "the student's anxiety decreased when you acknowledged their feelings". This targeted feedback loop is far more informative than the kind of generalized feedback one might get in a real classroom observation, and it accelerates skill acquisition in the same way a flight simulator allows a pilot to practice crisis maneuvers repeatedly in a short time.

\subsection{Metacognitive Skill Development}
Finally, using the system can enhance metacognition for both learners and instructors. Metacognition—thinking about one's own thinking and learning processes \cite{Flavell1979}—is crucial for self-improvement. By analyzing the simulation’s internal decision process, teacher trainees learn to think about the causes of student behavior in a structured way. Over time, they may internalize this habit, becoming more adept at diagnosing real students’ difficulties (e.g., recognizing signs of low self-efficacy or high anxiety in the classroom) and reflecting on their own teaching strategies.

Moreover, the explicit modeling of psychological constructs encourages a form of metacognitive awareness in instructors regarding their pedagogical content knowledge: it prompts them to consider, for each lesson, not just the content, but how student motivation, prior knowledge, and emotions might interplay during learning. In other words, it provides a scaffold for thinking about thinking—both the student’s and the teacher’s.

\section{Conclusion}
We have presented a novel multi-agent architecture for simulating human behavior that integrates AI technology with foundational psychological theory. By decomposing a virtual persona into multiple interacting psychological agents, the system achieves a level of realism and transparency that is difficult to obtain with conventional AI or simplistic role-play. The simulation can model domain-specific anxieties, fluctuating motivation, and interpersonal differences in a way that aligns with decades of research in psychology and education.

For educators and trainers, this offers a powerful new tool to practice and refine skills in handling complex human behaviors—bridging the longstanding gap between theory and practice. For researchers, it provides an experimental platform to test and visualize theoretical constructs in action. Ultimately, the work demonstrates how bridging AI and psychology can yield not only more effective training technologies but also deeper insights into the human mind, by literally making the invisible dynamics of thought and emotion visible and interactive.


\printbibliography
\end{document}